\documentclass{iise}

\usepackage{multirow}
\usepackage{subcaption} 
\conference{Proceedings of the IISE Annual Conference \& Expo 2025 \\
E. Gentry, F. Ju, X. Liu, eds.}

\title{\titlesize A TWO-LEVEL ENSEMBLE APPROACH FOR DIABETIC RETINOPATHY PREDICTION}
\author{
\normalfont Mahyar Mahmoudi$^1$, Tieming Liu$^1$\\
$^1$\normalfont Oklahoma State University, Stillwater, OK, USA
}

\abstractID{8562}

\begin{document}
\maketitle

\begin{abstract}
{\small Diabetic retinopathy (DR) is one of the leading causes of blindness in the working-age population. Current diagnostic methods for DR, comprehensive ophthalmic exams, are constrained by the availability of ophthalmologists and special equipment. Retinal-image-based AI tools for DR screening are not effective in detecting DR in early stages. We propose a novel non-image-based, two-level ensemble predictive model to identify patients with DR using routine lab test results. In the foundational stage of the ensemble, each base model—Linear SVC, Random Forest, Gradient Boosting, and XGBoost—is optimized through hyperparameter tuning and undergoes an internal stacking process across different configurations. This initial stacking step involves configuring each model to maximize individual scoring metrics, such as accuracy, precision, recall, and ROC-AUC, thus achieving optimal performance on a balanced set of evaluation criteria. A second level ensemble aggregates these base model initial predictions with another stacking step using Random Forest as the meta-learner to form the final predictive model. Compared with deep learning models, this two-level ensemble offers explainability and computational efficiency, ensuring interpretability in clinical contexts, while also achieving better performance. The two-level ensemble approach provides both diversity among the base models and eradicates the potential drawbacks of single-model approaches, making ensemble models a strategic choice to achieve high accuracy and adaptability to diverse clinical data. The model achieved a performance of Accuracy 0.9433, F1 Score 0.9425, Recall 0.9207, Precision 0.9653, ROC-AUC 0.9844, and AUPRC 0.9875, surpassing existing models in DR prediction.}
\end{abstract}
\section*{Keywords}
Ensemble Modeling, Diabetic Retinopathy Prediction, Stacking Models, Healthcare Data Analytics, Multi-Level Optimization

\section{Introduction}
The field of machine learning (ML) has increased in significance as an effective approach to predictive modeling for healthcare applications \cite{nithya2017predictive}. Traditional predictive models use one algorithm and are thus not very effective at coping with complex medical data \cite{datta2022hyper,zhao2019predicting}. Ensemble learning, particularly stacking, has emerged as an effective strategy to enhance predictive accuracy by combining the strengths of multiple ML models. However, most existing ensemble models use simple averaging or single-layer stacking \cite{shorewala2021early}, leaving room for improvement in model generalization and interoperability. This paper presents a two-level ensemble learning approach with the aim of obtaining optimal predictive performance in a computationally effective way. 

We test the performance of this method on a healthcare dataset for DR prediction from routine laboratory test results. DR is used as a case study application to demonstrate the ability of our method in a real-world classification problem. DR represents one of the most common causes of vision loss and blindness in diabetic patients \cite{ogurtsova2017idf}. Its prevalence keeps growing worldwide, and early diagnosis and intervention are crucial to prevent irreversible damage of the retina \cite{selvachandran2023developments}, yet current methods of screening for DR remain resource-intensive, relying on specialized equipment and trained ophthalmologists \cite{jaffe2004optical}. While there have been deep learning models for image-based DR detection, these often remain limited by data availability and high computational costs prohibitive in large-scale early screening \cite{balyen2019promising}. Predictive models using routine laboratory tests can therefore be an interesting alternative that can allow early identification of at-risk patients without specialized imaging.

\section{Related Research}
Ensemble learning has been widely applied to healthcare predictive modeling in order to improve classification accuracy and generalization \cite{an2023comprehensive}. Traditional ensemble methods, such as bagging and boosting, improve predictive stability but do not fully leverage the complementary strengths of diverse models \cite{nahar2019comparative}. Stack Generalization \cite{wolpert1992stacked} known as Stacking, a more advanced technique, integrates multiple classifiers by training a meta-learner on their outputs \cite{sesmero2015generating,rostami2024crispr}, leading to improved model robustness \cite{nahar2019comparative}.  A few studies illustrated stacking for medical diagnosis tasks. Hasan et al. \cite{hasan2020diabetes} proposed a stacking ensemble for diabetes prediction, which combined decision trees, random forests, and XGBoost to achieve higher accuracy than using a single classifier. Piri et al. \cite{piri2017data} implemented a stacking model for DR classification, underlining that the selection of base models significantly influences final performance.
These works thus identify that ensemble learning requires well-optimized base models, which aligns with the structure of our proposed two-level framework.

Furthermore, feature selection is important to enhance the interpretability of the model, along with its computational efficiency \cite{cheng2024comprehensive,li2017feature}. Wang et al. \cite{wang2021derivation} showed how the reduction in feature dimensionality in ensemble models for DR detection significantly improved their recall and precision. Similarly, Datta et al. \cite{datta2022hyper} and Nahar et al. \cite{nahar2019comparative} proposed that feature selection helps in alleviating overfitting in structured healthcare datasets of ensemble models through the elimination of irrelevant features. Our study further confirms these findings; the proposed two-level ensemble model expresses high recall and precision with only 6–7 features, therefore being computationally sufficient for clinical deployment.

On the other hand, Deep learning models, have been of great attention in healthcare analytics because they are useful for the capturing of complicated patterns of data \cite{miotto2018deep,shamshirband2021review}. However, their effectiveness depends on large datasets and high computational resources \cite{chen2014big}. Gadekallu et al. \cite{gadekallu2023deep} compared FCNs against ensemble models for DR prediction, concluding that while deep networks require by far longer training times, they cannot generalize well to limited-feature datasets. Ayon and Islam \cite{ayon2019diabetes} investigated Deep Neural Networks (DNNs) for the prediction of diabetes, showing that though DNNs give very good performance in applications where big data is available, having a black-box nature and relying on very large feature vectors renders them impractical. Sharma and Parmar \cite{sharma2020heart} also concluded that deep learning models do not behave well in small, structured datasets-a scenario often encountered in medical diagnosis. This is further supported by our findings, where the proposed ensemble model outperforms FCN in terms of predictive accuracy and computational efficiency using fewer features.


While there has been a great amount of success with ensemble stacking, prior research has not fully explored the combination of internal stacking at the level of base models with a hierarchical two-level stacking structure. Most research was either basic model stacking approaches or shallow ensembling techniques that did not allow further fine-tuning of the model predictions. This paper aims to fill this gap by proposing a two-level ensemble model that is far more accurate, less dependent on features, and computationally more efficient. The study illustrates the ability of internal stacking, meta-learning, and feature selection to make ensembles a scalable and interpretable alternative to deep learning for structured medical data classification.
\section{Methodology}
\subsection{Data Collection and Preprocessing}
This research leverages data from the Cerner's Health Facts database, a strong electronic health record system covering patient demographics, comorbidities, and laboratory test results. This data contains both DR patients and those without the disease, thus a suitable dataset for predictive modeling.

In order to ensure the quality of the data, variables with more than 20\% missing values were removed, and remaining missing values were imputed using mean values. The dataset was highly imbalanced, containing a lot more non-DR cases than DR cases. Thus, the Synthetic Minority Over-sampling Technique was performed in order to make a balanced class distribution by generating synthetic samples.

Retaining 25 key variables, including laboratory test results, demographic attributes, and comorbidity indicators, the dataset was then split into training and testing sets using an 80-20 ratio. Before model training, all numerical features were standardized to ensure uniform scaling across variables.
\subsection{Two-Level Ensemble Learning Framework}
This study introduces a two-level ensemble learning framework that is designed to enhance predictive accuracy and robustness without compromising computational efficiency. Conventional ensemble methods, such as bagging and boosting, combine predictions from multiple models but fail to exploit the strengths of individual classifiers fully. In contrast, stacking enhances performance by learning an optimal combination of base model predictions through a meta-learner.  While stacking has been used in many applications, most of the implementations have only a single level, and thus their ability to refine is bounded. We therefore now investigate the integration of multiple models hierarchically by a novel two-level stacking approach, combining not just individual model strengths but also ensemble diversity.

First of all, four base models at the first level of the ensemble are optimized with hyperparameter tuning and internally stacked: Random Forest, Gradient Boosting, Extreme Gradient Boosting, and Linear Support Vector Classifier. In other words, numerous configurations of each model have to be trained with the aim of maximizing the performance metrics for each one. Instead of selecting a single optimal configuration per model, multiple top-performing versions based on accuracy, precision, and recall are combined using stacking within each base model. This internal ensembling enhances the stability and robustness of each model before they contribute to the final prediction. The output of each base model is defined and showed as:

\begin{equation} \hat{y}i = g{\text{stack}}\left( h_1^i(X), h_2^i(X), ..., h_m^i(X) \right) \label{eq1} \end{equation}
where \( X \) represents the input features, \( h_1^i, h_2^i, ..., h_m^i \) are the multiple tuned versions of the \( i \)-th base model, and \( g_{\text{stack}} \) is the internal stacking function that aggregates these outputs. The result, \( \hat{y}_i \), is the optimized output of the \( i \)-th base model before second-level stacking.

\begin{figure}[htb]
	\centering
	\includegraphics[width=5.0in]{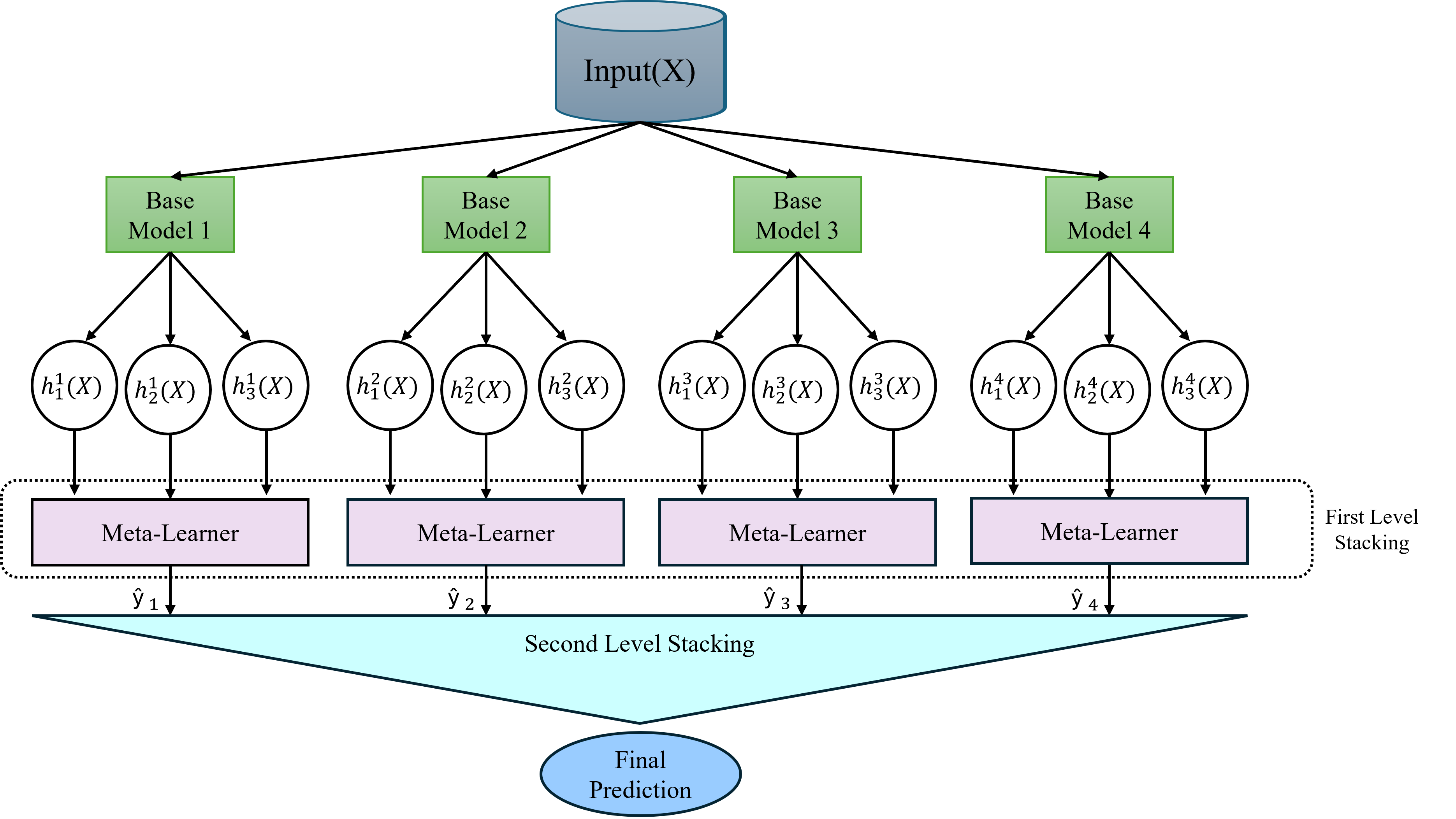}
	\caption{Flow diagram of two-level ensembling approach}\label{fig1}
\end{figure}

Then, the second level of the framework aggregates the outputs of the optimized first-level models using another stacking step, where a meta-learner refines predictions based on their combined outputs. Random Forest(RF) and Logistic Regression(LR) were evaluated as potential meta-learners, with RF ultimately chosen for its ability due to its ability to efficiently capture non-linear relationships and interactions between model outputs. The second level is defined as:
\begin{equation}
\hat{y} = f_{\text{meta}}\left( \hat{y}_1, \hat{y}_2, ..., \hat{y}_n \right)
\label{eq2}
\end{equation}
where \(\hat{y}_1, \hat{y}_2, ..., \hat{y}_n\) are the optimized first-level outputs of base models, and \( f_{\text{meta}} \) is the second-level stacking function applied to their predictions. This hierarchical ensembling approach reduces model bias, improves generalization, and enhances predictive performance as shown in Figure \ref{fig1}.

\section{Results \& Discussion}
\subsection{Performance of the Two-Level Ensemble Model}
The two-level ensemble model proposed in this paper has shown very good performance in DR classification. Among all the tested configurations, the two best models had the best accuracy of 0.9433, with a best F1 score of 0.9425. In the case of recall and precision for DR prediction, it is essential; therefore, for both, the model showed a very good performance: recall was 0.9214 while precision was 0.9646. These are further confirmed by the AUC-ROC and AUPRC values of the ensemble, with values of 0.9844 and 0.9875, respectively. A comparison of different configurations is shown in Table \ref{tab:performance_metrics}.
\begin{table*}[htb]
\centering
\caption{Performance Metrics of proposed model, one level stacking, and FCN}
\label{tab:performance_metrics}
\vspace{-0.3cm}
\begin{tabular}{l|cccccc}
\hline
Model  & Accuracy & F1 Score & Recall & Precision & ROC-AUC & AUPRC \\ \hline
LR-LR  & 0.9389  & 0.9387  & 0.9281  & 0.9496  & 0.9802  & 0.9831 \\ 
RF-LR  & 0.9397  & 0.9392  & 0.9243  & 0.9547  & 0.9819  & 0.9854 \\ 
RF-RF  & \textbf{0.9433}  & \textbf{0.9425}  & 0.9207  & \textbf{0.9653}  & \textbf{0.9844}  & \textbf{0.9875} \\ 
LR-RF  & \textbf{0.9433}  & \textbf{0.9425}  & \textbf{0.9214}  & 0.9646  & \textbf{0.9844}  & \textbf{0.9875} \\ 
\hline
One-Level Stacking  & 0.8958  & 0.8946  & 0.8769  & 0.9131  & 0.9650  & 0.9650 \\ 
FCN  & 0.8300  & 0.8200  & 0.8200  & 0.8400  & 0.8900  & 0.8800 \\ 
\hline
\end{tabular}
\end{table*}

\subsection{Comparison with One-Level Stacking and FCN}
For the one-level Stacking model, the accuracy was 0.8958, the F1 score was 0.8946, and the recall was 0.8769, which is lower as compared to the two-level ensemble. The two-level approach has always outperformed the one-level model by refining the predictions through the hierarchical stacking mechanism.

The FCN model tuned using four hidden layers with 256-128-64-32 neurons and ReLU activations resulted in an accuracy of 0.83 with an F1 score of 0.82. Though FCN achieved a recall of 0.82, FCN did not perform well in precision at 0.84, yielding more false positives than the proposed ensemble model. Also, FCN training took 200 minutes compared to 150 minutes required by the two-level ensemble, which can be computed using the same machine, proving the efficiency of the proposed model. The comparison of deep learning model and one-level stacking has been shown in Table \ref{tab:performance_metrics}. These results clearly show that the proposed two-level ensemble model consistently outperforms the one-level stacking and FCN on accuracy, recall, and precision, while being superior in computational efficiency.
\subsection{Impact of Feature Selection}
To evaluate the robustness of the proposed two-level ensemble model, we tested its performance using a varying number of features with RF in both stacking levels, starting with the top five most important features and incrementally increasing up to all 25 features. Feature importance was determined using permutation, which measures the performance drop when each feature is randomly shuffled. This approach ensures that selected features contribute meaningfully to the final ensemble predictions. The results indicate that with only 6–7 features, recall and precision remained close to the full-feature model. However, the performance drop from 6 to 5 features was more significant than from 7 to 6, reinforcing the necessity of using at least six features. This is particularly significant in clinical contexts, where reduced input numbers are associated with quickened decision-making and reduced computation. In direct contrast, the FCN performed best when exposed to all of the features of the dataset-a clear indication of its reliance upon a larger size of feature space for effective training. Table \ref{tab:feature_performance} presents a comparison of those results against those from the model with a complete 25-feature set.

\begin{table*}[htb]
\centering
\caption{Performance of Two-Level Ensemble Model with Different Feature Sets}
\label{tab:feature_performance}
\vspace{-0.3cm}
\begin{tabular}{l|cccccc}
\hline
Feature Count  & Accuracy    & F1 Score  & Recall  & Precision  & ROC-AUC  & AUPRC \\ \hline
5 Features     & 0.8900      & 0.8888    & 0.8600  & 0.9195     & 0.9550   & 0.9549 \\ 
6 Features     & 0.9175      & 0.9162    & 0.8809  & 0.9515     & 0.9662   & 0.9752 \\ 
7 Features     & 0.9212      & 0.9201    & 0.8895  & 0.9555     & 0.9718   & 0.9765 \\ 
25 Features    & 0.9433      & 0.9425    & 0.9207  & 0.9653     & 0.9844   & 0.9875 \\ \hline
\end{tabular}
\end{table*}

The top six most important features, ranked based on feature importance scores from the two-level ensemble model, include \textbf{Creatinine Serum Quantitative}, \textbf{Blood Urea Nitrogen}, \textbf{Anion Gap Blood}, \textbf{Glucose Serum Plasma Quantitative}, \textbf{Albumin Serum}, and \textbf{Alanine Aminotransferase}. These features encompass key renal, metabolic, and hematological markers, reinforcing previous clinical studies on DR risk factors \cite{piri2017data,homayouni2022diabetic}. Their significance aligns with existing medical research, particularly highlighting the role of kidney function, blood glucose regulation, and liver enzymes in DR progression.


Furthermore, feature selection experiments show that the performance of the ensemble model degrades gracefully with as few as 6–7 features, confirming scalability for a real-world application. This, in turn, underlines the efficiency and practicality of the proposed method compared to deep learning models, which cannot afford reduced feature sets. These findings may indicate that the two-level ensemble can serve as a more efficient alternative to deep learning models, balancing predictive power with computational efficiency.
\section{Conclusions}
This paper proposed a two-level ensemble learning framework that can improve the predictive accuracy and computational efficiency of structured medical data classification. The proposed model with hierarchical stacking outperforms benchmark deep learning models while retaining computational feasibility. Its ability to deliver high recall and precision even in reduced dimensions points toward its potential resource-efficient deployment in clinical applications.

Consequently, the findings of this research will position the ensemble-based methods as an effective, scalable alternative to deep learning models for healthcare predictive modeling, particularly in the context of small data availability. Directions for future research can be the improvement of model interpretability and robustness on diverse datasets. These entail the application of SHAP-based explanations to provide transparent, patient-level explanation of predictions and the assessment of model performance on subgroups or external cohorts to establish generalizability. While availability of large-scale behavioral or socioeconomic data may remain limited in the near term, the model can be improved further by leveraging more structured and available features—e.g., medication history or visit patterns—to minimize biases. Furthermore, this two-level ensemble approach can be extended to the early detection of other chronic conditions such as cardiovascular disease or kidney failure, so it becomes more applicable in practice.

\section*{Acknowledgements}
The research is supported by NIH through Project 5R01EY033861, entitled Harnessing Tensor Information to Improve EHR Data Quality for Accurate Data-driven Screening of Diabetic Retinopathy with Routine Lab Results.


\begin{thebibliography}{1}



\bibitem{nithya2017predictive}
Nithya, B., and Ilango, V., 2017, \newblock ``Predictive Analytics in Health Care Using Machine Learning Tools and Techniques," \newblock
Proc. of the 2017 International Conference on Intelligent Computing and Control Systems (ICICCS), IEEE, 492--499.


\bibitem{zhao2019predicting}
Zhao, J., Gu, S., and McDermaid, A., 2019, \newblock ``Predicting Outcomes of Chronic Kidney Disease from EMR Data Based on Random Forest Regression," \newblock
Mathematical Biosciences, 310, 24--30.

\bibitem{datta2022hyper}
Datta, P., Das, P., and Kumar, A., 2022, \newblock ``Hyperparameter Tuning-Based Gradient Boosting Algorithm for Detection of Diabetic Retinopathy: An Analytical Review," \newblock
Bulletin of Electrical Engineering and Informatics, 11(2), 814--824.

\bibitem{shorewala2021early}
Shorewala, V., 2021, \newblock ``Early Detection of Coronary Heart Disease Using Ensemble Techniques," \newblock
Informatics in Medicine Unlocked, 26, 100655.

\bibitem{ogurtsova2017idf}
Ogurtsova, K., da Rocha Fernandes, J. D., Huang, Y., Linnenkamp, U., Guariguata, L., Cho, N. H., Cavan, D., Shaw, J. E., and Makaroff, L. E., 2017, \newblock ``IDF Diabetes Atlas: Global Estimates for the Prevalence of Diabetes for 2015 and 2040," \newblock
Diabetes Research and Clinical Practice, 128, 40--50.

\bibitem{selvachandran2023developments}
Selvachandran, G., Quek, S. G., Paramesran, R., Ding, W., and Son, L. H., 2023, \newblock ``Developments in the Detection of Diabetic Retinopathy: A State-of-the-Art Review of Computer-Aided Diagnosis and Machine Learning Methods," \newblock
Artificial Intelligence Review, 56(2), 915--964.

\bibitem{jaffe2004optical}
Jaffe, G. J., and Caprioli, J., 2004, \newblock ``Optical Coherence Tomography to Detect and Manage Retinal Disease and Glaucoma," \newblock
American Journal of Ophthalmology, 137(1), 156--169.

\bibitem{balyen2019promising}
Balyen, L., and Peto, T., 2019, \newblock ``Promising Artificial Intelligence-Machine Learning-Deep Learning Algorithms in Ophthalmology," \newblock
The Asia-Pacific Journal of Ophthalmology, 8(3), 264--272.



\bibitem{an2023comprehensive}
An, Q., Rahman, S., Zhou, J., and Kang, J. J., 2023, \newblock ``A Comprehensive Review on Machine Learning in Healthcare Industry: Classification, Restrictions, Opportunities and Challenges," \newblock
Sensors, 23(9), 4178.





\bibitem{nahar2019comparative}
Nahar, N., Ara, F., Neloy, M. A. I., Barua, V., Hossain, M. S., and Andersson, K., 2019, \newblock ``A Comparative Analysis of the Ensemble Method for Liver Disease Prediction," \newblock
Proc. of the 2nd International Conference on Innovation in Engineering and Technology (ICIET), IEEE, 1--6.






\bibitem{wolpert1992stacked}
Wolpert, D. H., 1992, \newblock ``Stacked Generalization," \newblock
Neural Networks, 5(2), 241--259.

\bibitem{sesmero2015generating}
Sesmero, M. P., Ledezma, A. I., and Sanchis, A., 2015, \newblock ``Generating Ensembles of Heterogeneous Classifiers Using Stacked Generalization," \newblock
Wiley Interdisciplinary Reviews: Data Mining and Knowledge Discovery, 5(1), 21--34.

\bibitem{rostami2024crispr}
Rostami, M., Ghariyazi, A., Dashti, H., Rohban, M. H., and Rabiee, H. R., 2024, \newblock ``CRISPR: Ensemble Model," \newblock
arXiv preprint arXiv:2403.03018.



\bibitem{hasan2020diabetes}
Hasan, M. K., Alam, M. A., Das, D., Hossain, E., and Hasan, M., 2020, \newblock ``Diabetes Prediction Using Ensembling of Different Machine Learning Classifiers," \newblock
IEEE Access, 8, 76516--76531.

\bibitem{piri2017data}
Piri, S., Delen, D., Liu, T., and Zolbanin, H. M., 2017, \newblock ``A Data Analytics Approach to Building a Clinical Decision Support System for Diabetic Retinopathy: Developing and Deploying a Model Ensemble," \newblock
Decision Support Systems, 101, 12--27.

\bibitem{cheng2024comprehensive}
Cheng, X., 2024, \newblock ``A Comprehensive Study of Feature Selection Techniques in Machine Learning Models," \newblock
(Unpublished or Missing Journal Information).

\bibitem{li2017feature}
Li, J., Cheng, K., Wang, S., Morstatter, F., Trevino, R. P., Tang, J., and Liu, H., 2017, \newblock ``Feature Selection: A Data Perspective," \newblock
ACM Computing Surveys (CSUR), 50(6), 1--45.

\bibitem{wang2021derivation}
Wang, R., Miao, Z., Liu, T., Liu, M., Grdinovac, K., Song, X., Liang, Y., Delen, D., and Paiva, W., 2021, \newblock ``Derivation and Validation of Essential Predictors and Risk Index for Early Detection of Diabetic Retinopathy Using Electronic Health Records," \newblock
Journal of Clinical Medicine, 10(7), 1473.



\bibitem{miotto2018deep}
Miotto, R., Wang, F., Wang, S., Jiang, X., and Dudley, J. T., 2018, \newblock ``Deep Learning for Healthcare: Review, Opportunities and Challenges," \newblock
Briefings in Bioinformatics, 19(6), 1236--1246.

\bibitem{shamshirband2021review}
Shamshirband, S., Fathi, M., Dehzangi, A., Chronopoulos, A. T., and Alinejad-Rokny, H., 2021, \newblock ``A Review on Deep Learning Approaches in Healthcare Systems: Taxonomies, Challenges, and Open Issues," \newblock
Journal of Biomedical Informatics, 113, 103627.

\bibitem{chen2014big}
Chen, X.-W., and Lin, X., 2014, \newblock ``Big Data Deep Learning: Challenges and Perspectives," \newblock
IEEE Access, 2, 514--525.

\bibitem{gadekallu2023deep}
Gadekallu, T. R., Khare, N., Bhattacharya, S., Singh, S., Maddikunta, P. K. R., and Srivastava, G., 2023, \newblock ``Deep Neural Networks to Predict Diabetic Retinopathy," \newblock
Journal of Ambient Intelligence and Humanized Computing, Springer, 1--14.

\bibitem{ayon2019diabetes}
Ayon, S. I., and Islam, M. M., 2019, \newblock ``Diabetes Prediction: A Deep Learning Approach," \newblock
International Journal of Information Engineering and Electronic Business, 13(2), 21.

\bibitem{sharma2020heart}
Sharma, S., and Parmar, M., 2020, \newblock ``Heart Diseases Prediction Using Deep Learning Neural Network Model," \newblock
International Journal of Innovative Technology and Exploring Engineering (IJITEE), 9(3), 2244--2248.

\bibitem{peerbasha2023diabetes}
Peerbasha, S., Iqbal, Y. M., Praveen, K. P., Surputheen, M. M., and Raja, A. S., 2023, \newblock ``Diabetes Prediction Using Decision Tree, Random Forest, Support Vector Machine, K-Nearest Neighbors, Logistic Regression Classifiers," \newblock
Journal of Advanced Applied Scientific Research, 5(4), 42--54.

\bibitem{kumari2021ensemble}
Kumari, S., Kumar, D., and Mittal, M., 2021, \newblock ``An Ensemble Approach for Classification and Prediction of Diabetes Mellitus Using Soft Voting Classifier," \newblock
International Journal of Cognitive Computing in Engineering, 2, 40--46.

\bibitem{homayouni2022diabetic}
Homayouni, A., Liu, T., and Thieu, T., 2022, \newblock ``Diabetic Retinopathy Prediction Using Progressive Ablation Feature Selection: A Comprehensive Classifier Evaluation," \newblock
Smart Health, 26, 100343.


\end{thebibliography}
\end{document}